\documentclass{article}
\usepackage{arxiv}

\usepackage[utf8]{inputenc} % allow utf-8 input
\usepackage[T1]{fontenc}    % use 8-bit T1 fonts
\usepackage{hyperref}       % hyperlinks
\usepackage{url}            % simple URL typesetting
\usepackage{booktabs}       % professional-quality tables
\usepackage{amsfonts}       % blackboard math symbols
\usepackage{nicefrac}       % compact symbols for 1/2, etc.
\usepackage{microtype}      % microtypography
\usepackage{graphicx}
\usepackage{natbib}
\usepackage{doi}

\usepackage{lineno,hyperref}
\modulolinenumbers[5]

\usepackage{graphicx}
\usepackage{comment}
\usepackage{color} 
\usepackage{xcolor}
\usepackage[utf8]{inputenc}

\definecolor{mypink1}{rgb}{0.858, 0.188, 0.478}
\definecolor{mypink2}{RGB}{219, 48, 122}
\definecolor{mypink3}{cmyk}{0, 0.7808, 0.4429, 0.1412}
\definecolor{mygray}{gray}{0.6}
\definecolor{grey}{rgb}{0.7, 0.75, 0.71} 
\definecolor{almond}{rgb}{0.94, 0.6, 0.4}

\definecolor{Mycolor2}{HTML}{00F9DE}

\usepackage{varwidth}
\usepackage{tabularx}
\usepackage{caption}
\title{Whose Opinions Matter? Perspective-aware Models to Identify Opinions of Hate Speech Victims in Abusive Language Detection}

\date{} 					% Or removing it

\author{
    Sohail Akhtar\\
	Department of Computer Science\\
	University of Turin\\
	Turin, Italy\\
	\texttt{sohail.akhtar@unito.it}\\
	\And
	Valerio Basile \\
	Department of Computer Science\\
	University of Turin\\
	Turin, Italy\\
	\texttt{valerio.basile@unito.it}\\
	\And
	Viviana Patti \\
	Department of Computer Science\\
	University of Turin\\
	Turin, Italy\\
	\texttt{viviana.patti@unito.it} \\
}

\renewcommand{\shorttitle}{}

\hypersetup{
pdftitle={Whose Opinions Matter? Perspective-aware Models to Identify Opinions of Hate Speech Victims in Abusive Language Detection},
pdfsubject={q-bio.NC, q-bio.QM},
pdfauthor={Sohail Akhtar, Valerio Basile, Viviana Patti},
pdfkeywords={Abusive language detection, linguistic annotation, perspectives identification, inter-rater agreement, data augmentation, social media},
}

\begin{document}
\maketitle

\begin{abstract}

Social media platforms provide users the freedom of expression and a medium to exchange information and express diverse opinions. Unfortunately, this has also resulted in the growth of abusive content with the purpose of discriminating people and targeting the most vulnerable communities such as immigrants, LGBT, Muslims, Jews and women. Because abusive language is subjective in nature, there might be highly polarizing topics or events involved in the annotation of abusive contents such as hate speech (HS). Therefore, we need novel approaches to model conflicting perspectives and opinions coming from people with different personal and demographic backgrounds.
In this paper, we present an in-depth study to model polarized opinions coming from different communities under the hypothesis that similar characteristics (ethnicity, social background, culture etc.) can influence the perspectives of annotators on a certain phenomenon. We believe that by relying on this information, we can divide the annotators into groups sharing similar perspectives. We can create separate gold standards, one for each group, to train state-of-the-art deep learning models. We can employ an ensemble approach to combine the perspective-aware classifiers from different groups to an inclusive model. We also propose a novel resource,
a multi-perspective English language dataset annotated according to different sub-categories relevant for characterising online abuse: hate speech, aggressiveness, offensiveness and stereotype. Unlike our previous work, where the annotations were based on crowd-sourcing, here, we involved the victims of targeted communities in the annotation process, who volunteered to annotate the dataset, providing a natural selection of the annotator groups based on their personal characteristics. By training state-of-the-art deep learning models on this novel resource, we show how our approach improves the prediction performance of a state-of-the-art supervised classifier.
Moreover, we also perform an in-depth qualitative analysis of the novel dataset to identify and understand the relevant keywords, topics and events causing polarization among the annotators in expressed opinions. 

\end{abstract}

% keywords can be removed

\section{Introduction}

Hate speech (HS) is a special type of abusive language whose detection on social media platforms is a rather difficult but important task. \textit{Online hate speech} (or \textit{cyber-hate}) may take different forms. The sudden rise in hate speech related incidents on social media is considered a major issue, especially targeting vulnerable communities such as immigrants, Muslims, women and LGBT+ people~\citep{Duggan2017}. The factors often responsible for such hate may include the demographic background and personal characteristics of the victims such as ethnicity, religion, race, sexual orientation and color ~\citep{Nobata:2016:ALD:2872427.2883062,Duggan2017}. This online growth of HS is often considered a reason for violent behaviour and hatred towards the vulnerable communities~\citep{Izsak2015}. This issue needs to be addressed at a global level to counter the growing hate. The laws defined by some countries against abusive language consider certain expressions hateful and consequently illegal ~\citep{abbondante2017ruolo}. Judging whether a message contains hate speech is quite subjective, given the nature of the phenomenon. Hate speech online is not necessarily toned in an aggressive or offensive way. Rather, it is characterized by an explicit call to violent action~\citep{Poletto2017HateSA}. It has been observed that this increase in online hate may result in violent acts against people belonging to different communities, bringing serious social consequences~\citep{articlemossie19,OKeeffe2011TheIO}. 

In recent years, hand in hand with the massive rise of online hatred~\citep{articleZhang}, the research community has shown great interest in HS detection and classification, also due to considerable social impact of the phenomenon on the well-being of a society~\citep{jurgens2019just}. Hate speech is a highly subjective phenomenon and an extremely complex notion.
In online discussions, especially on the controversial topics, the users have shown a stronger tendency to express their opinions~\citep{mikeWendling}, and this resulted in the spread of hate against people.

Hate speech is also a highly controversial topic. Controversy in social media texts stem from events, topics or social issues that generate different responses from online users~\citep{Popescu:2010:DCE:1871437.1871751}. High controversy is bound to have an impact on the manual annotation of abusive phenomenon, in terms of agreement between the human judges. Since the manual annotation of a public dataset is a crucial step in creation of language resources that are used for training predictive models of a language, the controversial text might cause performance issues for Natural Language Processing (NLP) approaches that rely on such supervised machine learning techniques. Most abusive language detection corpora are composed of data collected from social media platforms~\citep{Poletto2020ResourcesAB}, such as Twitter and Facebook. Most of them are collected by querying social media APIs with specific lists of keywords. Then, the data are annotated by human judges either by relying on crowd-sourcing platforms or on experts (often judges with knowledge of the subject). The annotated datasets are then employed to detect opinions expressed by online users for predefined categories such as presence vs. absence of a specific phenomenon (e.g., offensive behaviour, hate speech against immigrants, cyber-bullying, and so on). When the annotation process relies on crowd-sourcing, in most cases, such platforms do not provide any background information (culture, ethnicity, social background etc.) on the workers annotating the datasets. However, even if this information is available, the cultural background of the annotators is usually considered a secondary aspect or only a single culture is preferred for annotation~\citep{inproceedings12345Sheerman11}. 

\cite{sap-etal-2019-risk} highlighted the problem of racial bias in existing approaches for online abusive language detection tasks and provided an empirical characterization of such bias prevalent in social media platforms. They emphasized that there is a strong relationship between AAE markers (e.g, “n*ggas”,“ass”, "f*ck") and HS annotations and the models trained on such annotated data are highly likely to pick up and replicate this bias in the data. Such problems with the datasets may raise ecological and methodological issues especially when studying phenomena that target specific communities such as immigrants. This a limit in general, and especially in the case of HS detection. Indeed, HS against different communities often contains stereotypical words which might arouse positive or negative sentiments or different reactions in annotators with different backgrounds~\citep{Warner:2012:DHS:2390374.2390377, waseem2016you}. Due to constant unavailability of information on annotators, there are high chances of observing polarization among the judgements on potentially same abusive messages, which might result in low inter-annotator agreements. The problems with the quality of gold standard data when dealing with subjective phenomena have been investigated recently, e.g., by~\cite{articleVBasile}, where the manual annotation of subjective phenomena is found to be tainted by serious issues in terms of inter-annotator agreements. As a consequence, the benchmarks based on the datasets created with traditional methods are found to be inadequate and leading to unstable results.

Researchers, who recently started tackling hate speech detection from a natural language processing perspective. are designing operational frameworks for HS, annotating corpora with several semantic frameworks, and automatic classifiers based on supervised machine learning models~\citep{article3fortuna18,inproceedings10schmidt17,poletto2019annotating}. Most of the classifiers proposed in literature for hate speech detection are based on supervised machine learning approaches, which require large corpora annotated by humans~\citep{BPB13} and the development of gold standards and the benchmarks to train machine learning models for specific classification tasks and compare the results of different systems. 

In order to measure the agreement between annotators, Fleiss’ Kappa and Krippendorff’s Alpha are commonly used by NLP research community to assess the quality of data annotations. However, such agreement measures are not always free of shortcomings~\citep{Artstein2008SurveyAI}. In particular, the issues with these commonly used agreement measures for datasets annotated by the workers from crowd-sourcing platforms have been recently investigated by~\cite{CheccoRMMD17}, which also provided solutions to solve these problems with alternative measures. Similarly,~\cite{hovy-etal-2013-learning} proposed a method called MACE to address the issue of low reliability of crowd-sourced annotators and its impact on the quality of gold standard produced with such methods. However, when we consider the case of HS datasets, there are also further task specific relevant issues that need to be considered when we reason about different perspectives and disagreement among the annotators on the presence of hate. This is especially true in toxic online environments, where discussion often turns into abusive expressions of hate against vulnerable groups, such as immigrants, LGBTQ+ people and women. Ideally, involving the victims and targets of hate speech in the annotation process would help us to understand their views on hate related incidents. However, the fine-grained information about the cultural, ethnicity, or demographic background of the annotators is usually not available, or it is not a primary factor when selecting experts or volunteers as annotators. We therefore propose a methodology to automatically model the different perspectives that annotators may adopt towards certain highly subjective phenomena, i.e., abusive language and hate speech. In our method, supervised machine learning models are trained to learn different point of view of the human annotators on same data.

We propose a comprehensive and novel computational study to investigate a more robust conceptualization of different types of abusive phenomena such as hate speech, aggressiveness, offensiveness and also related phenomena such as stereotype. We provide a deeper analysis of a novel dataset with a natural division of annotators into homogeneous groups with the aim of improving automatic detection of online abusive language by taking into account the single opinions of annotators and how they diverge on certain topics. Our working hypothesis is that the difference in opinions expressed by different groups of annotators is a valuable source of information rather than a noise factor in the creation of a gold standard dataset. By processing such information, we aim at creating better quality data to train abusive language models for the prediction of highly subjective phenomenon such as cyber-hate. In particular, we focus on inter-annotator agreements computed for subdivisions of the annotator set, and on the level of polarization of annotated texts. We experimented with state-of-the-art transformer-based Neural models, such as BERT~\citep{DBLP:journals/corr/abs-1810-04805}, which is a pre-trained language model fine-tuned for each down-stream detection task.

This work extends and complete preliminary studies of the authors~\citep{10.1007/978-3-030-35166-3_41, Akhtar_Basile_Patti_2020}, by providing further analysis and additional experiments. 
In previous work, we initially experimented on three available datasets in English and Italian, featuring different topical focuses. In this paper, we further extend our work by developing a novel multi-perspective dataset of abusive language. This dataset is annotated manually by a group of annotators having different personal, cultural and demographic background, including migrants as the victims of online abuse and providing a deeper insight and better understanding of our hypothesis regarding the importance of capturing annotators' perspectives and developing perspective-aware models.

\subsection{Research Questions}
\label{subsec:RQuest}

In this study, we address the following research questions:

\textbf{RQ1:} \textit{How can we measure the level of polarization among the annotators when their judgements reflect different perspectives in an experimental setup where either cultural and demographic backgrounds of the annotators are available or we do not have any information about the annotators?}
%\noindent

In order to answer the question, we developed a novel approach based on our previous work~\citep{10.1007/978-3-030-35166-3_41}, a polarization index (P-index) which measures the level of polarization (personalized perspectives) expressed by annotators in opinions given on individual instances of an abusive phenomenon when the annotated data is crowd-sourced and in general, we do not have any information on the background of annotators. The P-index is a valuable tool to aid the division of annotators into groups by performing an exhaustive search and finding the partition having maximum polarization. We tested it on several available benchmark datasets. In this study, we do not need P-index to aid the division of annotators into similar groups as we did for the datasets annotated by crowd-sourcing workers because we have a natural selection of the annotator groups based on their personal characteristics.

\textbf{RQ2:} \textit{Can a measure of polarization for individual instances of subjective content help us exploring the datasets and understanding the topics and issues involved with polarizing nature?}

In order to answer this qualitative research question, we will employ P-index with the newly developed multi-perspective abusive language dataset. With the help of P-index, the most polarized messages naturally emerge at the top of the list. Similarly, we can analyze the group-wise predictions to find the topics and keywords relevant to certain events which cause polarization among the annotators.

\textbf{RQ3:} \textit{Can we improve the classification performance of machine learning models by introducing training sets with different perspectives and is it possible to effectively represent these perspectives expressed by annotators in an inclusive model?}

To answer this question, we create separate gold standard for a dataset for each of the annotator groups and perform classification tasks to measure the performance of perspective-aware supervised models in which the model learns from the training sets created from different groups. We also employed an ensemble classifier, which was proposed in previous work, that considers all the learned perspectives in an inclusive fashion.

\subsection{Contributions}
\label{subsec:Contb}

The summary of our contributions in this study are given below:

\textbf{1.} We present an overview of the state-of-the-art in abusive language detection.

\textbf{2.} We provide a detailed review of the novel approaches developed in our previous work which include a polarization index, the unique concept of the creation of group-based gold standards, and the development of an inclusive model to detect different forms of abusive language in social media with state-of-the-art NLP models~\citep{10.1007/978-3-030-35166-3_41, Akhtar_Basile_Patti_2020}.

\textbf{3.} We developed a multi-perspective abusive language dataset containing tweets downloaded from Twitter. The dataset is manually labelled and the complete information about the personal, cultural and demographic backgrounds of the annotators is available. For the first time, we involved migrants as the victims of abusive language to annotate the dataset. This helped us to capture the common perspectives that may shape their opinions. It is an important step in our study because we believe that the current datasets and annotation schemes available for abusive language detection fail to model and understand the feeling and emotions expressed by the victims of online and offline abuse.

\textbf{4.} We present a comprehensive and in-depth qualitative analysis of our dataset which provides a deep and thorough understanding of polarized instances and how they might affect the gold standard creation process and particularly, the annotation process. The qualitative analysis also helps us to explore and understand the topics, events and keywords related causing polarization among the annotators.

The rest of the article is organized as follows. Section~\ref{sec:RWork} presents the literature review on abusive language detection. Then, we introduce Section~\ref{sec:ABLDetection} which explains the details about the tasks and methodology. The details about the newly developed multi-perspective BREXIT dataset are explained in Section~\ref{sec:brexit}. The datasets from previous work are described in Section~\ref{sec:data}. We present the results of our empirical evaluation on the available datasets in Section~\ref{sec:evaluation}. Then we present the qualitative analysis performed by manual exploration of the datasets in Section~\ref{sec:Qanalysis}, before the conclusion in Section~\ref{sec:conclusion}. 

\section{Related Work}
\label{sec:RWork}

As a complex phenomenon, hate speech online typically concerns how various social media groups and communities develop a relationship while exploring social media platforms. Although there are several definitions of what constitutes hate, still there is no formal agreeable consensus on HS definition~\citep{inproceedings4ross16}. This has, unfortunately, made it difficult for state-of-the-art models to effectively combine and compare the performance of detection systems developed for different types of online abusive content (e.g, hate speech, aggressiveness, misogyny etc.) in social media datasets. Hate speech usually refers to disparaging individuals or groups because of their ethnicity, gender, race, color, sexual orientation, nationality, religion, or other similar characteristics, see for instance the U.S. constitution~\citep{Nockleby2000}. Several social media platforms such as Twitter~\footnote{\url{https://www.twitter.com/}} typically implement their own definitions of hate speech, when they define their policies on content moderation and terms of use. For instance, the definition by Twitter is: 
"Hateful conduct: You may not promote violence against or directly attack or threaten other people on the basis of race, ethnicity, origin, sexual orientation, gender, gender identity, religious affiliation, age, disability, or disease”.

The recent rise in the production of content on social media platforms and increase in the number of users have also seen a steady growth in hate speech~\citep{Wiegand2019DetectionOA}. Most of the current research on HS and abusive language detection tries to identify online hate by employing supervised learning approaches to detect and prevent it. For a general purpose HS detection task, the researchers use expert and amateur annotators to annotate the corpora downloaded from various social media sites such as Twitter and Facebook. \cite{article3fortuna18}, in their recent survey on HS literature, provided a deep and thorough analysis of issue and challenges of automatic HS detection faced by research community. They also addressed the issue of the availability of high quality datasets for bench-marking and training the models for HS detection tasks.

Recently, several campaigns were organized to evaluate the targeted hate in a multi-lingual perspective~\citep{hateval2019semeval,Fersini2018OverviewOT,DBLP:conf/evalita/BoscoDPST18,DBLP:conf/evalita/FersiniNR18}. The work by~\cite{Warner:2012:DHS:2390374.2390377} used the term hate speech and focused on collecting HS messages against Jews from various social media sites and classifying HS based on the stereotypical words commonly used in an antisemitic manner. \cite{Benesch2016} developed a system to detect HS in social media text and provide counter-narratives against such hate arising from certain political events. While there exists legislation governing hate speech, often social media platforms implement their own regulations. Social media has provided new opportunities to collect and analyze rich data on abusive language, including online harassment and hate against women and communities often targeted based on their ethnicity and gender~\citep{Duggan2017,waseem2016you}. 

Controversiality is not a new concept in the study of social media. Usually, the controversial topics and issues are identified and the user responses or opinions on these topics or issues are gathered for further analysis. The focus in Sentiment Analysis is on the words or texts in online topics or news items which are controversial in nature~\citep{Beelen:2017:DCO:3077136.3080723}. Therefore, the literature mostly focuses on the controversiality of a message or the other aspects of online content such as controversial events~\citep{Popescu:2010:DCE:1871437.1871751}, rather than the polarization of annotator's opinions. Some common controversial topics often discussed online include climate change, abortion, and vaccination among the others~\citep{DBLP:conf/clic-it/BasileCN17}. People from different communities and backgrounds react differently to controversial topics and sometimes, the discussions also make some topics controversial because of the presence of bias against a certain community~\citep{Beelen:2017:DCO:3077136.3080723,DBLP:conf/clic-it/BasileCN17}. The discussion on controversial topics often results in the spread of online hate~\citep{Popescu:2010:DCE:1871437.1871751}. The focus of studies, which measure the level of controversy by analyzing user opinions or responses on controversial topics, is on the controversy of topics or issues rather than on the level of polarization in opinions expressed by annotators when they annotate these datasets~\citep{Beelen:2017:DCO:3077136.3080723}. We may expect that highly controversial events, topics or issues can impact the agreement between annotators and this may lead to polarized judgements. 

Recent studies on inter-annotator agreement provide us an insight to the methodology and the effectiveness of annotation process. \cite{articleBhow08} used Kappa coefficients to measure the quality and reliability of effective human annotations and resulting corpora by classifying the single items into multiple categories. Gold standard datasets that are used for training models in NLP are traditionally created with manual annotation, whose quality is assessed by metrics of inter-rater agreement such as Fleiss’ Kappa~\citep{doi:10.1177/001316447303300309}. However, criticism has been raised towards such measures~\citep{inproceedingsPowers12}. These methods to measure the annotator agreements are not free from issues, especially when the datasets are annotated by crowd-sourcing workers~\citep{CheccoRMMD17}. The quality of gold standard datasets for subjective phenomena can be empirically tested, e.g., as in~\cite{articleVBasile}, where the authors proposed a set of experiments based on the agreements between different systems, expert annotators and the results of crowd-sourcing annotations. \cite{Soberon:2013:MCT:2874376.2874381} highlighted the importance of disagreements in a dataset and proposed a method to harness the disagreement of annotators as a source of knowledge rather than treating it as noise in the data. \cite{CheccoRMMD17} introduced new agreement metrics that aim to account for the polarization of annotator opinions. The quality of gold standard datasets can be tested empirically as in~\cite{basile2018sentiment}, where the authors compared experiments of agreement between different systems, expert annotators, and the results of crowd-sourcing annotations. The idea that the disagreement among the annotators is not a noise but resourceful knowledge to analyze the phenomenon to create better training sets is studied by~\cite{Soberon:2013:MCT:2874376.2874381}. In this study, we also leverage the disagreement among the annotators but we consider it as an interpretation of divergence of opinion between the groups.

Pre-trained language models have recently gained significant importance in the research community. These models rely on feature-based approaches where the structure of a down-stream task is used as an extra feature such as in ElMo~\citep{article1300peters18}. OpenAIGPT, a unidirectional model was proposed in~\cite{Radford2018ImprovingLU}, used a multi-layered transformer-based architecture with a left-to-right approach and fine-tuning with less dependency on the task-specific architectures. ULMFiT was proposed by~\cite{inproceedings2howard18} and intended for text classification achieving state-of-the-art results on several benchmark datasets. These models are unidirectional. This means while processing the tokens, they either work from left-to-right or from right-to-left, thus limiting the power of pre-trained language representations.

We proposed a method in~\cite{10.1007/978-3-030-35166-3_41} to improve text classification in a supervised setting. The tasks were typically framed as binary classification tasks, e.g., presence vs. absence of hate speech. In a supervised learning scenario, we need training sets labelled by human judges following specific guidelines and an annotation scheme. We proceeded with the intuition that a classifier can learn better if there is less polarization in opinions expressed by annotators, meaning that the value of polarization index is lower. Similarly, if the value of P-index is higher, this may cause confusion for the classifier resulting in lower performance. To test the method, the P-index is computed for individual instances and the messages are replicated based on the the computed P-index value. We only replicated the instances in the training set while no change was made to the test set. The instances were replicated a number of times inversely proportional to their computed P-index. Instances with maximum P-index value were deleted from the training set. We modified the training sets and kept the test sets fixed without any intervention. We employed a straightforward supervised classifier based on Support Vector Machine Model (SVM) with bag-of-words as features and employed TF-IDF as weighting strategy. We optimized unigrams as main features in the vectorized tweets. With these experiments, the classification performance generally improved for all the datasets over baseline values. Although, we obtained good results with the classification experiments, still there was a need to find the computational approaches to model the opinions expressed by annotators with a specific background while annotating a dataset.

Summarizing, to our knowledge, this paper is the only study in which we explored the concept of perspective-aware systems in combination with the development of a multi-perspective corpus annotated by migrants as the victims of online abuse and grouping the annotators having different social and demographic backgrounds. Previous work focused on abusive language in general or without considering the modeling of perspectives expressed in opinions by the annotators. Modeling polarized opinions and group based classifiers were explored in our work~\citep{10.1007/978-3-030-35166-3_41, Akhtar_Basile_Patti_2020}, however not focusing on the development and experimentation on a multi-perspective corpus.

\section{Automatic Abusive Language Detection: Task and Methodology}
\label{sec:ABLDetection}

This section provides a detailed description of the abusive language detection task. First, we continue with a brief description of the task, the goals we desire to achieve by performing the task, and then we explain the methodology applied to perform the task. Finally, we give details about the evaluation strategy.

\subsection{Task Description}
\label{subsec:TDescription}

Automatic abusive language detection is cast as a text classification task. The purpose of the task is to differentiate between hateful and non-hateful, aggressive and non-aggressive, offensive and non-offensive, and stereotype and non-stereotype content in a binary classification scenario, which can be featured by different topical focuses depending on the targets of hate. The main task is divided into many sub-tasks.

The sub-tasks are defined below:

\textbf{1.} Measurement of polarization index: The first sub-task is to measure an index that identifies the level of polarization in opinions given by annotators belonging to different groups (human judges) who annotated a given dataset. It is important to note that the polarization index was developed in our previous work and explained in methodology section.

\textbf{2.} Creating annotator groups: If we do not have any information on the annotators background, we measure polarization index for individual instances of a dataset, then we divide the annotators of that dataset into groups by performing an exhaustive search. The split that maximizes the average polarization index is selected for the division of annotators. In principle, If we have any information on the personal and demographic backgrounds of all annotators, we do not need to find a split by performing an exhaustive search as in such a case, we have a natural selection of annotator groups. In this study, we developed a multi-perspective dataset with manual annotation process in which, we have complete information on the annotators so we have a natural selection for grouping the annotators based on the available information.

\textbf{3.} Creating group-wise gold standards: We create separate gold standards for annotator groups. The idea is that each gold standard training data can represent individualized perspective expressed in the annotations by annotators belonging to that particular group. These perspective are modeled using machine learning algorithms for abusive language detection.

\textbf{4.} Classification tasks: Finally, we perform binary classification and evaluate the performance of group-wise classifiers. We also employ an ensemble approach which combines all the perspective-aware classifiers into a an inclusive model.

The classification performance is evaluated in terms of accuracy, precision, recall and f1 score. 

\subsection{Method}
\label{subsec:method}

We proposed a novel approach in~\cite{10.1007/978-3-030-35166-3_41} to exploit the fine granularity of single annotations of highly subjective phenomena resulting from crowd-sourcing platforms. The purpose was to create higher quality gold standards for supervised learning of a subjective phenomenon such as hate speech. We introduced a new index that measured the level of polarization in each instance of a dataset annotated by different annotator groups. We assumed that all the annotators who participated in the annotation process can be divided into sub-groups based on some common characteristics from various traits like race, social grouping, cultural background, education, sex, color, and other similar factors. We believed that common personal traits can shape the opinions of human judges while annotating textual instances. 

First, we showed that how traditional inter-annotator agreement metrics can provide new insight when applied in a setting, where the annotators do not form one homogeneous group. Then, we introduced a new index that measured the level of polarization of a message with respect to the annotations given by two different groups. We brought the results of these metrics into a supervised learning context by developing straightforward methods. These methods automatically manipulated a training set based on its single annotations. We measured the divergence of opinions between different combinations of annotators by using the newly developed index.

\subsection{Measurement of Polarization Index}
\label{subsec:PIndex}

Many datasets for NLP tasks are annotated by experts or crowd-sourced workers. Thus in general, the background of annotators is not known. However, we hypothesize that a group of annotators can be effectively divided according to the characteristics linked to their background, by analyzing their annotations. In particular, we make use of the polarization Index (P-index) introduced in~\cite{10.1007/978-3-030-35166-3_41} and its application, described in the same article, for dividing annotators based on the polarization of their judgments. This division can induce higher quality data for supervised learning tasks for subjective phenomena such as hate speech and abusive language. The method leverages the information at the level of single annotations, measuring the level of polarization for all the annotations individually. We believe that the knowledge about a topic, the background, and or social circumstances may generate polarized opinions among different communities. This may be reflected in the annotations given by the human judges from different communities on highly subjective phenomena. In this scenario, we are testing a \textit{homophily} hypothesis~\citep{article123mcpherson01} which states that social groups can strongly shape their social network. We postulate that in a similar way, the common background of people can also shape their opinions which results in polarized judgments if they are asked to annotate messages from a certain domain. The domain knowledge can also influence these opinions. 

It is important to note that there is a difference between the polarization of opinions and inter-annotator agreement as the latter might be influenced by factors such as the knowledge of a language, text comprehension and interpretation, e.g., of ironic content while the former stems from the level of subjectivity of some phenomena (e.g., hate speech is highly subjective). In our work, we capture the annotator background at a macro-level. It is interesting to note that when we have high polarization among the instances of dataset, it does not necessarily mean that there is low agreement between the annotator groups: according to our definition, we consider the judgements highly polarized when different groups have high agreement on different judgements. On the contrary, there is no polarization at all if we have an overall low agreement between different groups or among the members of same group.

The first step of the method consist of automatically dividing the set of annotators into groups. In particular, we compute the division that maximizes the average P-index, by performing an exhaustive search of the possible annotator partitions and computing the average P-index for each partition. In this study, we limited the possible partitions to two groups. 

Suppose we have a set of messages represented by $N$ and a set of annotators represented by $M$. Suppose that $g_{i,j}$ denotes the annotation of a human judge $j$ on a given message $i$. For each message $i \in N$, we can divide the set of its annotations $G_i=\{g_{i,1}, ... g_{i,m}\}$ into $k$ subsets $G^1_i$, ..., $G^k_i$.

Secondly, we measure the agreement between the annotator groups for annotations on single messages by using normalized $\chi^2$ statistics. These statistics test how independent the distribution of annotations are when measured against a uniform distribution. The idea is that there is a total disagreement between the annotators if the distribution of annotations is uniform. As an example, if three of the six annotators in a binary setting decide for one label, and remaining three chose a different label, we will have a (3,3) distribution of the labels which is uniform showing a maximum disagreement between the groups. When we normalize the $\chi^2$ by dividing the statistics by number of annotations, we may get a value between 0 (total disagreement) and 1 (perfect agreement):

\begin{equation}
\label{eq:agreement}
a(G_i) = 1 - \frac{\chi^2(G_i)}{|M|}
\end{equation}

\noindent
We compute the \textit{polarization} index (P-index) of a message $i$ as: 
\begin{equation}
\label{eq:cscore}
P(i)=\frac{1}{k}\sum_{1\leq w \leq k} a(G^w_i)  (1 - a(G_i))
\end{equation}

\noindent
The value of P is between 0 and 1, where 0 means that there is no polarization among the members of annotator groups on a single message whereas, 1 indicates that the level of polarization is maximum between the groups. It is important to note that the P-index differs from inter-annotator agreement in a sense that the value of P-index for a message is high if different groups show high agreement on different opinions. On the other hand, if overall agreement is low even between the members of same group, there is no polarization in the groups for that message.

To give a few examples with 
suppose, $k=2$, if $G^1_i = \{1, 0, 0\}$ and $G^2_i = \{1, 1, 1\}$ , then $a(G^1_i) \approx 0.11$ (low intra-group agreement), $a(G^2_i)=1$ (high intra-group agreement), $a(G_i) \approx 0.11$, thus $P(i) \approx 0.49$. 
If we have $G^1_i = \{0, 0, 0\}$ and $G^2_i = \{1, 1, 1\}$, which means that each group is in total agreement but on different labels, then $a(G^1_i)=1, a(G^2_i)=1$, $a(G_i) = 0$, thus $P(i) = 1$. 
Similarly, if there are two annotators in group 1 and three in group 2 (Homophobia dataset) then with $k=2$, if $G^1_i = \{0, 0\}$ and $G^2_i = \{1, 0, 1\}$ , then $a(G^1_i) =1$ (high intra-group agreement), $a(G^2_i) \approx 0.11$ (low intra-group agreement), $a(G_i) \approx 0.04$, thus $P(i) \approx 0.53$.

\subsection{Perspective-Aware System Modeling}
\label{subsec:PerspectiveMod}

In previous section, we introduced a method to create different annotator groups with the help of P-index. With the knowledge obtained by measuring the level of polarization and division of annotators, we can automatically model the perspectives coming from different annotators that may adopt towards certain highly subjective phenomena, i.e., abusive language and hate speech. The annotator groups are beneficial in terms of creating gold standards and developing models trained to learn different perspectives of human judges on similar data. 

As we assumed that the polarization index can successfully divide the annotators into groups based on their personal characteristics, hence the instances annotated by each group represent the opinions of people belonging to that group. We can create separate gold standards; one for each group based on majority voting method and train state-of-the-art classifiers on group-based training data. We believe that the models trained on a gold standard annotated by people with a common personal background can represent these personalized perspectives and also give us an insight how well the classifier performed with that particular group after the models are evaluated. We proposed perspective-aware models in our previous work~\citep{Akhtar_Basile_Patti_2020} Where we tested the methodology by performing classification experiments on different Twitter datasets in English and Italian languages, featuring different forms of hate speech: sexist, racist and homophobic contents. We also proposed an ensemble classifier approach that consider all the learned perspectives in an inclusive fashion. The ensemble classifier considers an instance positive if any of the group 1 or group 2 classifiers (or both) considers it positive. We call this ensemble ``Inclusive''. The rationale behind this ensemble is that hate speech is a sparse and subjective phenomenon, where each personal background induces a perspective that leads to different perceptions of what constitute hate. This classifier includes all these perspectives in its decision process. 

The Inclusive classifier will naturally have a bias towards the positive class, by construction. It is important to note that in previous work, we did not have a natural selection of the annotator groups and by utilizing P-index capabilities, we divided the annotators into groups. In this study, we have a natural selection of the annotator groups. 

\begin{figure*}[ht]
\begin{center}
    \centering
    \includegraphics[width=0.9\linewidth]{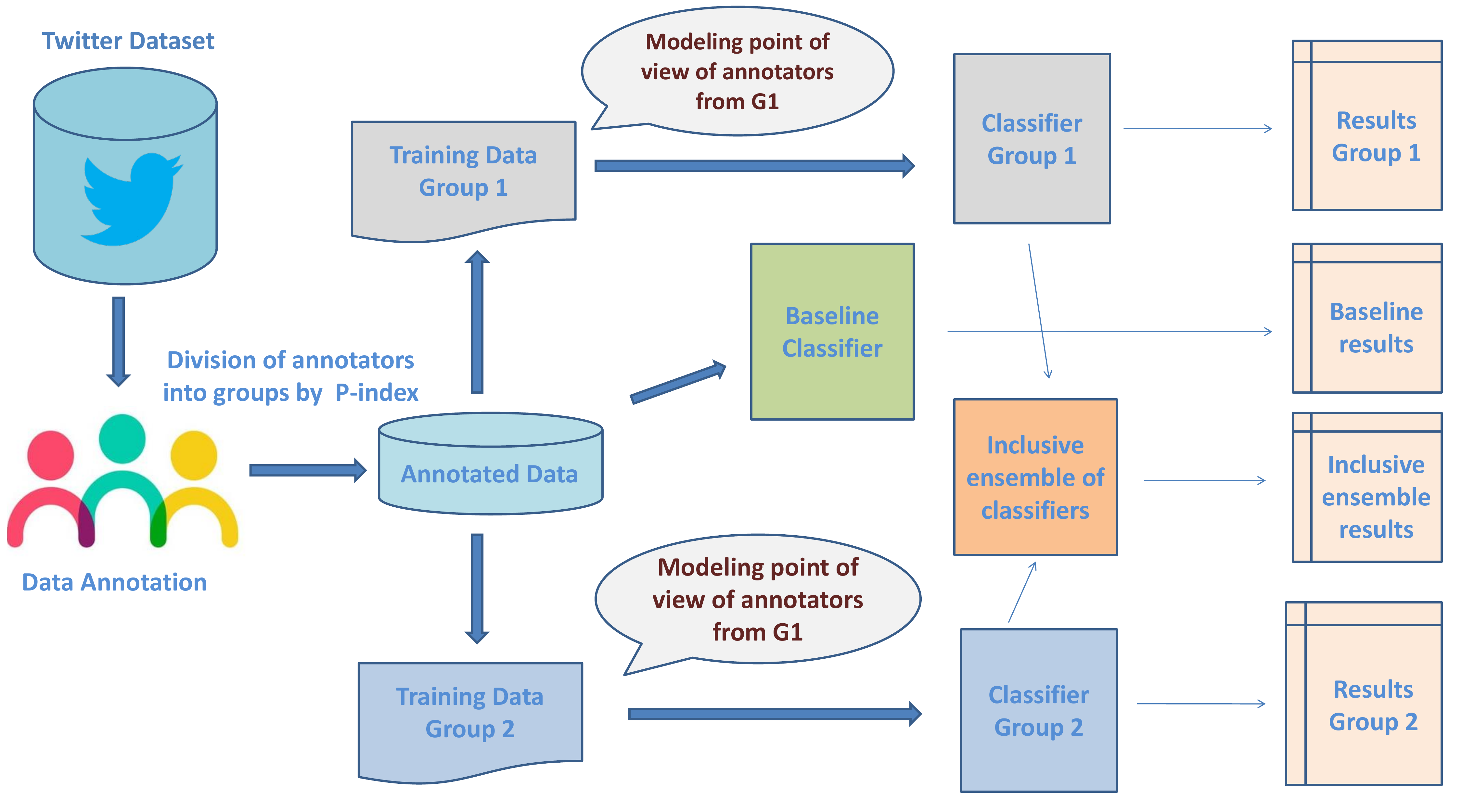}
    \caption{The architecture of proposed approach for perspective-aware abusive language detection.}
    \label{fig:architecture}
\end{center}
\end{figure*}

Figure~\ref{fig:architecture} explains the architecture of perspective-aware system to detect abusive language. 

\section{Development of a Multi-perspective Abusive Language Dataset on Brexit}
\label{sec:brexit}

Brexit is a portmanteau of the words "British" and "exit" merged to refer to the exit of the great Britain from the European Union (EU). Social media platforms are highly influential in encouraging users to express opinions on certain world level events, such as Brexit. There is an increased user participation to openly express their opinions and suggestions on such events. During the voting period of Brexit referendum, there was an exponential increase in the number of opinions expressed online, suggesting different reasons and expressing intent concerning the importance of the event. 

The data for current research was gathered from~\cite{articleLai19} in the context of a study concerning {\em stance detection} (referred as BREXIT dataset henceforth). Data were collected in the form of tweets for different time intervals from Twitter. The time frame for collection of data on Brexit was before the voting, during the voting, and immediately after the voting period for referendum to measure the frequency of tweets and the stance of online users about the event. 

The motivations to focus on the Brexit political debate are explained in the following. The process of creating an abusive language dataset with a step-by-step narrative requires the raw data to be rich enough for mining contents which are more resourceful and provide better insights about a topical phenomenon, in our case abusive language. The research community should be able to consider the approaches for transforming the existing insulting environment of social media into a non-hate inclusive online society. Because of its global importance and strong impact on European society, we decided to have a through review of the available BREXIT dataset, also stimulated by studies highlighting how the Brexit debate have been over-determined by racist and xenophobic attitudes~\citep{miller2016brussels,racismandbrexit20}. Careful manual analysis of the tweets revealed interesting patterns from the discussions on topics and events strongly linked with abusive contents including racism. Majority of the tweets blamed immigrants and Muslims as reasons for the Brexit and also using derogatory and abusive words against minorities. 
This motivated us to select the dataset for our research work on abusive language detection.

\subsection{Data Collection and Filtering}
\label{subsec:DFProcessing}

Initially, around 5 million tweets were collected between June $22^{nd}$ and $30^{th}$, 2016 by using the hashtag \#Brexit. After getting the dataset, we performed many pre-processing steps to filter and clean the dataset for further work. We divided the collected data into sub-corpora related to three categories: Immigration, Islamophobia, and Xenophobia. The details about different categories are given below. For each category, keywords have been selected based on a previous study~\citep{miller2016brussels}. The following list shows the keywords used to filter the dataset: \textit{Immigration, migration, immigrant, migrant, foreign, foreigners, terrorism, terrorist, Muslim, Islam, jihad, Quran, illegals, deport, anti-immigrant, rapefugee, rapugee, paki, pakis, nigger}. Table~\ref{tab:keywords} shows the frequency of the occurrence of these keywords in the whole dataset.

\begin{table}[ht]
\caption{\label{tab:keywords} The frequency of keywords in the dataset.}
\setlength{\tabcolsep}{20pt}
\begin{center}
%\small
\begin{tabular}{rrr}
 \hline
 \textbf{Immigration}  & \textbf{Islamophobia} & \textbf{Xenophobia}\\
\hline
immigration (43287)    & Jihad (1974)          & Illegals (823) \\
migration (50365)      & Jihadi (1140)         & deport (5003) \\
immigrant (23803)      & terrorist (5161)      & rapefugee (550) \\
migrant (48334)        & terrorism (3765)      & rapeugee (41) \\
foreign (23661)        & muzzie (243)          & anti-immigrant (1776)  \\
foreigner (5793)       & muzzies(42)           & anti-immigration (1490)  \\
foreigners (5148)      & islam (30239)         & paki (8117)  \\
refugee (18519)        & kuffar (46)           & pakis (7692) \\
refugees (14416)       & kaffir (9)            & nigger (183) \\
                       & Quran (878)           &               \\
                       & muslim (26556)        &                \\
 \hline
\end{tabular}
\end{center}
\end{table}

\textbf{Immigration}:\\
We filtered the BREXIT dataset by using different keywords to select the tweets related to immigrants, and to gather and analyze the opinions expressed by online users about the immigration and the role of immigrants in a society. The list of keywords for immigration corpus is shown in Table~\ref{tab:keywords}. We retrieved a total of 53,824 tweets by filtering the data with related keywords.

\textbf{Xenophobia}:\\
We further filtered the dataset by using keywords that are linked to Xenophobia. The idea was to get a subset of the BREXIT corpus to analyze whether the data also contain xenophobic contents, which refer to hate against minorities and less privileged communities in a society. We retrieved a total of 4585 tweets by using keywords that are mentioned in Table~\ref{tab:keywords} with the frequency of occurrence of these keywords in the dataset.

\textbf{Islamophobia}:\\ 
Similar to Immigration and Xenophobia corpora, we filtered the BREXIT corpus by using keywords specific to Islamophobia to get a subset of the data. The details about the keywords and their occurrence in the overall dataset are mentioned in Table~\ref{tab:keywords}. We retrieved a total of 17222 tweets by using keywords related to Islamophobic content.

\textbf{Discarded racism and racist keywords}:\\
When reasoning on the relevant keywords to select our data, we also investigated the occurrence of two important keywords, {\em racist} and {\em racism}, in order to determine their use by tweeters of the BREXIT dataset in various contexts. The frequency of the occurrence of keyword ‘racist’ in the whole data set was 72,023, which shows a relatively high usage frequency. Similarly, the word ‘racism’ occurred 54,024 times which also shows a high usage of the keyword. However, a close manual analysis of the tweets where these keywords occur revealed that most of the tweets were targeting natives of the Britain as racist people or were linking racism to the reason why Britain is leaving the EU. Some of them  highlighted that the racism in the UK is on rise, other tweets also included abusive language targeting people living in the UK and voting for Brexit.

Therefore, we came to the conclusion that most of the tweets filtered using such keywords were not containing abusive language against immigrants, but were directed towards the natives of the UK. Therefore, we decided not to further use {\em racism} and {\em racist} as keywords to select data. We decided to report here about the process that led us to consider and discard these keywords, to bring to the light some further aspects on the nature of the original dataset.

\begin{figure*}[ht]
\begin{center}
    \centering
    \captionsetup{justification=centering}
    \includegraphics[width=0.9\linewidth]{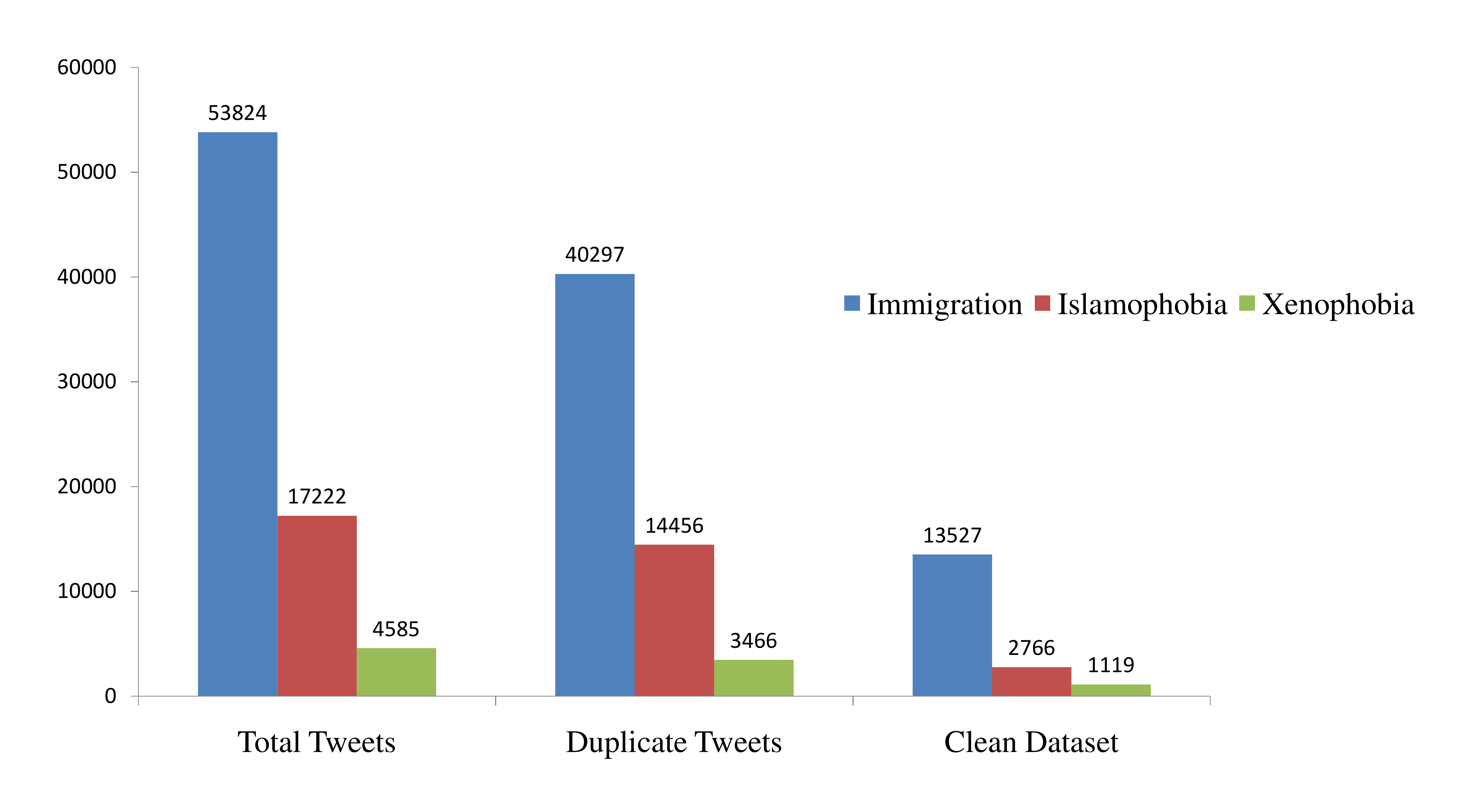}
    \caption{Details of the clean sub-sections of the dataset}
    \label{fig:untitled}
\end{center}
\end{figure*}

\textbf{Pre-processing}:\\
After selecting the data as specified before, we applied pre-processing steps before the annotation experiments. We cleaned the data set by removing duplicates including the retweets.
Figure~\ref{fig:untitled} shows a comparison of three sub-corpora with the number of retrieved tweets.

\subsection{Data Annotation}
\label{subsec:annotations}

The next important step is the data annotation process. One of the focus of this study is to involve the victims of abusive language such as immigrants and Muslims in the annotation process. To create an annotated corpus, we randomly selected 1120 tweets from the dataset. The corpus was annotated according to four categories that include \textit{Hate Speech}, \textit{Aggressiveness}, \textit{Offensiveness}, and \textit{Stereotype}, following the  multi-layered annotation scheme suggested by~\cite{Sangui-LRECL18-1443} to develop a corpus of hate speech against immigrants. A total of six annotators were selected for the annotation process and then divided into two groups. Since the categories are subjective in nature, we thought it would be interesting to see the results of annotation experiments and agreement measures between the two groups. 

The annotation process for different types of abusive language is a rather difficult and vague process which usually results in low agreement scores as also acknowledged in~\cite{inproceedings10schmidt17}. Here, the annotators were briefed and trained so that they have a similar understanding of the abusive categories. 

The scheme and guidelines proposed are described and referred in~\cite{Poletto2017HateSA,Sangui-LRECL18-1443}. The authors used the scheme and guidelines to annotate an Italian language dataset on HS against Muslims and Roma. The guidelines were written in Italian language. We translated them into English and modified them according to our requirements to educate the annotators with minor modifications. All the annotators were volunteers with certain demographic and cultural background. The first group of the three volunteers were first-or second-generation immigrants and students from developing countries to Europe and the UK, of Muslim background. The other group has three volunteers who were researchers with western background and having experience in linguistic annotation. The same dataset was annotated by two groups and also followed the same scheme of annotation and guidelines. We named the groups as \textit{Target} which are the migrants and \textit{Control} which are the locals. For further processing, we will also call the \textit{Control} group as \textit{Group 1} and the \textit{Target} group as \textit{Group 2}. 
In order to mitigate the possible gender bias, we also involved people with different gender in each group of annotators.

\subsection{Dataset Statistics}
\label{subsec:stats}

In this section, we will give a statistical description and quantitative analysis of the final dataset with all sub-categories. We can see the distribution of labels for all categories in Table~\ref{tab:data12}. It is clear from the distribution that the abusive language categories are highly unbalanced. This means that there are less instances of positive class and more of the negative class in our data. Although, for the Offensiveness and Stereotype, the ratio of positive class is relatively better in comparison with the Hate Speech and Aggressiveness categories.

\begin{table}[ht]
\caption{\label{tab:data12} BREXIT sub-categories with the distribution of labels.}
\setlength{\tabcolsep}{20pt}
\begin{center}
\begin{tabular}{lccr}
\hline
\textbf{Dataset} &\textbf{Positive class} & \textbf{Negative class} & \textbf{Total} \\
\hline
Hate Speech    & 106 & 1,014 & 1,120\\  
Aggressiveness & 87  & 1,033 & 1,120\\ 
Offensiveness  & 206 & 914   & 1,120\\
Stereotype     & 151 & 969   & 1,120\\
\hline
\end{tabular}
\end{center}
\end{table}

Table~\ref{tab:agrMeasure} shows the average P-index measured for all the possible splits by performing an exhaustive search for each of the BREXIT category. In order to validate the metric, we need an annotated dataset in which the personal and demographic backgrounds of the annotators are known and the annotators having similar background are grouped together. In order to measure the average, we first calculated the P-index values of individual sentences in a category and then we measured the average value of P-index for all the instances in each of the categories. If the average is higher, we have more polarization and divergence of opinions among the annotators for that category. 

As shown in Table~\ref{tab:agrMeasure}, the value is higher for HS, Aggressiveness and Offensive for the natural grouping of annotators which indicate how effectively the P-index can divide the annotators coming from different communities and backgrounds into groups based on the divergence of their opinions. This Table also shows the maximum and minimum values for all the other possible splits which are less than the values of natural grouping. It is interesting to see that for Stereotype, the average P-index value of another split is slightly higher than the natural split. One possible reason is that we found high pairwise agreements between one member of the Control group and all other members of the Target group and this might cause the average p-index value to go slightly higher than the natural grouping.

It is important to note that we conducted a similar study in our previous research~\citep{10.1007/978-3-030-35166-3_41} but, the study was only limited to HS category and only a few tweets were available for the analysis. Here, we extended the abusive language categories and also the number of tweets for each category to further verify the validation process of the division of annotators into groups by using P-index.

\begin{table}[ht]
\caption{\label{tab:agrMeasure} BREXIT categories with natural grouping and maximum (other splits) and minimum (other splits) average P-index values.}
\setlength{\tabcolsep}{15pt}
\begin{center}
\begin{tabular}{lccc}
 \hline
  \textbf{Category} & \textbf{Natural} & \textbf{Max.(other splits)} & \textbf{Min.(other splits)}\\
 \hline
 Hate Speech    &\textbf{0.14}   &0.09           &0.06\\
 Aggressiveness &\textbf{0.10}   &0.09           &0.07\\
 Offensiveness  &\textbf{0.18}   &0.14           &0.11\\
 Stereotype     &0.14            &\textbf{0.15}  &0.12\\
 \hline
\end{tabular}
\end{center}
\end{table}

We measured different types of agreements among the annotators from the two groups. Table~\ref{tab:OverallAg} shows intra-group and overall agreements. We measured the agreements by using Fleiss’ kappa coefficient. As seen in Table~\ref{tab:OverallAg}, the value of overall agreement for each category is low when compared to the agreements between the members of individual groups. Also, the agreements for Hate Speech and Offensiveness are relatively higher than the other categories. We hypothesize that if the level of subjectivity of a task is higher, the value of kappa is low and vice versa. This means that Stereotype is the most subjective category in the dataset having very low kappa values because stereotypes are often more implicit in nature.

\begin{table}[ht]
\caption{\label{tab:OverallAg} Group-wise and overall agreements for all the categories.}
\setlength{\tabcolsep}{15pt}
\begin{center}
\begin{tabular}{lcccc}
 \hline
 \textbf{Agreements} &\textbf{HS} &\textbf{Aggress.}  &\textbf{Offens.} & \textbf{Stereotype}\\
 \hline
 Overall Agreement &0.35 &0.30 &0.36 &0.29\\
 Control Group         &0.43 &0.34 &0.44 &0.28\\
 Target Group         &0.58 &0.37 &0.49 &0.33\\
 \hline
\end{tabular}
\end{center}
\end{table}

The computation of pairwise agreements between the members of individual groups and between the members of two groups provided us a fine-grained network of agreements between all the annotators. Such topology of the network indicates the relationships between the opinions of annotators within a group and also between the groups. Tables~\ref{tab:pairwise1},~\ref{tab:pairwise2},~\ref{tab:pairwise3} and~\ref{tab:pairwise4} show us the pairwise agreements in an explainable manner. 

We can observe that for each of the categories in BREXIT dataset, the pairwise agreements between the members of same groups are rather higher than the agreements between the members of different groups. The value for HS category for the members of Control group is between 0.41 and 0.44 and between 0.52 and 0.66 in the Target group. However, the the value between the members of two groups is between 0.20 and 0.28 which is significantly low. Similarly, for Aggressiveness, the value between the pairs of same group is between 0.26 and 0.44 for the Control group and between 0.29 and 0.48 for the Target group which is high but the value between the pairs of different groups is in between 0.17 and 0.34 which is lower than the higher value between the pairs of same group. For Offensiveness, we see a similar pattern and the value is between 0.39 and 0.49 for the Control and between 0.30 and 0.36 for the Target group, a relatively higher value but between the pairs of different groups, the value is between 0.27 and 0.35 which is still lower than the other groups.
For Stereotype, we saw a slightly different pattern in which the pairwise agreements between one member of the Control group and all the members of the Target group are higher than all other pairwise agreements (three values in the top left row of Table~\ref{tab:pairwise4}). The pairwise agreements between the members of the Target group (between 0.30 and 0.36) were relatively higher than the Control group (between 0.23 and 0.37).

We can still deduce from the pairwise agreements that the two groups of annotators show much higher intra-group agreements (top-left and bottom-right area of the Tables) than their inter-group agreements (top-right area).

\begin{table}[ht]
\caption{\label{tab:pairwise1} Pairwise agreements between two groups for Hate Speech.}
\setlength{\tabcolsep}{12pt}
\centering
\begin{center}
\begin{tabular}{lccccc}
 \hline
      &  \color{almond}{C2} &  \color{almond}{C3} & \color{almond}{T1}  & \color{almond}{T2}  & \color{almond}{T3}\\                      
                                                 
 \color{almond}{C1}   &\colorbox{orange}{0.41} &\colorbox{orange}{0.45} &\colorbox{pink}{0.22} &\colorbox{pink}{0.20} &\colorbox{pink}{0.28}\\                   
 \color{almond}{C2}   &     &\colorbox{orange}{0.44} &\colorbox{pink}{0.22} &\colorbox{pink}{0.20} &\colorbox{pink}{0.23}\\                    
 \color{almond}{C3}   &     &     &\colorbox{pink}{0.28} &\colorbox{pink}{0.25} &\colorbox{pink}{0.27}\\                    
 \color{almond}{T1}   &     &     &     &\colorbox{yellow}{0.66 }&\colorbox{yellow}{0.56}\\                    
 \color{almond}{T2}   &     &     &     &     &\colorbox{yellow}{0.52}\\                    
\end{tabular}
\end{center}
\end{table}

\begin{table}[ht]
\caption{\label{tab:pairwise2} Pairwise agreements between two groups for Aggressiveness.}
\setlength{\tabcolsep}{12pt}
\centering
\begin{center}
\begin{tabular}{lccccc}
 \hline
      &  \color{almond}{C2} &  \color{almond}{C3} & \color{almond}{T1}  & \color{almond}{T2}  & \color{almond}{T3}\\
 %\hline
\color{almond}{C1}   &\colorbox{orange}{0.44} &\colorbox{orange}{0.33} &\colorbox{pink}{0.34} &\colorbox{pink}{0.27} &\colorbox{pink}{0.27}\\
 \color{almond}{C2}   &     &\colorbox{orange}{0.26} &\colorbox{pink}{0.24} &\colorbox{pink}{0.17} &\colorbox{pink}{0.31}\\
 \color{almond}{C3}   &     &     &\colorbox{pink}{0.27} &\colorbox{pink}{0.25} &\colorbox{pink}{0.24}\\
 \color{almond}{T1}   &     &     &     &\colorbox{yellow}{0.48} &\colorbox{yellow}{0.29}\\
 \color{almond}{T2}   &     &     &     &     &\colorbox{yellow}{0.31}\\
 \hline
\end{tabular}
\end{center}
\end{table}

\begin{table}[ht]
\caption{\label{tab:pairwise3} Pairwise agreements between two groups for Offensiveness.}
\setlength{\tabcolsep}{12pt}
\centering
\begin{center}
\begin{tabular}{lccccc}
 \hline
      &  \color{almond}{C2} &  \color{almond}{C3} & \color{almond}{T1}  & \color{almond}{T2}  & \color{almond}{T3}\\                         
 %\hline                                                
 \color{almond}{C1}   &\colorbox{orange}{0.49} &\colorbox{orange}{0.39} &\colorbox{pink}{0.35} &\colorbox{pink}{0.28} &\colorbox{pink}{0.38}\\ 
 \color{almond}{C2}   &     &\colorbox{orange}{0.40} &\colorbox{pink}{0.34} &\colorbox{pink}{0.28} &\colorbox{pink}{0.32}\\                 
 \color{almond}{C3}   &     &     &\colorbox{pink}{0.27} &\colorbox{pink}{0.24} &\colorbox{pink}{0.33}\\                   
 \color{almond}{T1}   &     &     &     &\colorbox{yellow}{0.55} &\colorbox{yellow}{0.49}\\                   
 \color{almond}{T2}   &     &     &     &     &\colorbox{yellow}{0.43}\\          
 \hline
\end{tabular}
\end{center}
\end{table}

\begin{table}[ht]
\caption{\label{tab:pairwise4} Pairwise agreements between two groups for Stereotype.}
\setlength{\tabcolsep}{12pt}
\centering
\begin{center}
\begin{tabular}{lccccc}
 \hline
     &  \color{almond}{C2} &  \color{almond}{C3} & \color{almond}{T1}  & \color{almond}{T2}  & \color{almond}{T3}\\
 %\hline                                                 
 \color{almond}{C1}   &\colorbox{orange}{0.37} &\colorbox{orange}{0.23} &\colorbox{pink}{0.41} &\colorbox{pink}{0.43} &\colorbox{pink}{0.41}\\ 
 \color{almond}{C2}   &     &\colorbox{orange}{0.25} &\colorbox{pink}{0.32} &\colorbox{pink}{0.29} &\colorbox{pink}{0.19}\\
 \color{almond}{C3}   &     &     &\colorbox{pink}{0.15} &\colorbox{pink}{0.15} &\colorbox{pink}{0.12}\\
 \color{almond}{T1}   &     &     &     &\colorbox{yellow}{0.36} &\colorbox{yellow}{0.30}\\
 \color{almond}{T2}   &     &     &     &     &\colorbox{yellow}{0.34}\\
 \hline
\end{tabular}
\end{center}
\end{table}

\section{Targeted Hate Speech in English and Italian Datasets}
\label{sec:data}
In this section, we will give an overview of the datasets we employed in our previous work to give a comparative analysis with the experiments in this study. The given information include details of individual data sets and the information about the background of annotators, if available. 

The first data set that was gathered from social media is about hate speech in English language. The dataset consists of tweets and was developed by~\cite{waseem2016you}. It is available on a Github repository\footnote{\url{https://github.com/ZeerakW/hatespeech}} and contains a total of 6,909 tweets. The authors of dataset only provided the Twitter IDs and labels. With the available IDs, when the tweets were queried on twitter, we only managed to get a smaller dataset consisting of 6,361 tweets because some of the tweets perished with the passage of time.

The original dataset was annotated as a multi-labelled task by experts (feminist and anti-racism activists) and non-expert workers hired via a crowd-sourcing platform \footnote{\url{https://www.figure-eight.com/}} with the annotation guidelines described in~\cite{N16-2013waseem}. While annotating the data, if the experts were not sure of any instance, they were allowed to skip those tweets. By majority voting method, the annotations from experts were aggregated into a single label. The workers were given only those tweets which were labelled by the experts. There were at least four annotators who annotated each tweet in the dataset. Due to privacy concerns, we could not find more details about the total number of annotators and the information about their personal background. With the available annotations, we computed the gold labels by majority voting. Since there were four annotators, if there was a tie between the annotators, the preference was given to the judgment of expert annotators for the gold label. All the annotators (experts and non-experts) were treated equally.

The hate speech dataset from~\cite{waseem2016you} was a multi-labelled set with four annotation categories: \textit{sexism}, \textit{racism}, \textit{both}, and \textit{neither}. In order to convert the dataset into a binary classification task, the sexism and racism classes were separated. The details are described below (Sections~\ref{subsec:sexism} and \ref{subsec:racism}).

To serve a multi-lingual perspective, an additional dataset of homophobic tweets in Italian language was employed. The Homophobia dataset consists of 1,859 tweets on hateful contents against LGBT community (see Section~\ref{subsec:homophobia}). The details of these datasets from previous studies with the distribution of labels are shown in Table~\ref{tab:data}. 

\begin{table}[ht]
\caption{Datasets with the distribution of labels.}
\label{tab:data}
\setlength{\tabcolsep}{16pt}
\begin{center}
\begin{tabular}{lccr}
\hline
\textbf{Dataset} &\textbf{Positive class} & \textbf{Negative class} & \textbf{Total} \\
\hline
Sexism & 810 & 5,551 & 6,361 \\  
Racism & 100 & 6,261 &  6,361 \\ 
Homophobia & 224 & 1,635 & 1,859 \\
\hline
\end{tabular}
\end{center}
\end{table}

\subsection{Sexism}
\label{subsec:sexism}
In Sexism dataset, the labels \textit{sexism} and \textit{both} were converted to \textit{sexist}, and similarly, the labels \textit{racism} and \textit{neither} were converted to \textit{non-sexist}. After the conversions, the {\sf Sexism} dataset contained 810 tweets out of 6,361 (12.7\%) which were marked as \textit{sexist}.

The inter-annotator agreement among the annotators and the overall agreement (Fleiss' Kappa) were measured for the Sexism dataset. The agreement among the annotators was 0.58, which indicated a relatively moderate agreement. The annotators were divided into two groups by following the methodology of annotator partition explained in the method section. The intra-group agreements between the members of two groups were 0.53 and 0.64 respectively.

\subsection{Racism}
\label{subsec:racism}
By following the same procedure applied to the Sexism dataset (Section~\ref{subsec:sexism}), the binary labelled {\sf Racism} dataset was derived from the data of~\cite{waseem2016you}. The same annotation scheme and guidelines were applied as with the original dataset explained in Section~\ref{sec:data}. 
However, the mapping scheme of original labels was different as here the labels: \textit{racism} and \textit{both} were mapped to \textit{racist}, whereas the labels \textit{sexism} and \textit{neither} were mapped to \textit{non-racist}. The resulting Racism dataset was composed of a total of 100 tweets out of 6,361 (1.57\%) which were marked as \textit{racist}.
 
The overall agreement (Fleiss' Kappa) was 0.23, which is relatively high disagreement among the annotators. The annotators of the Racism dataset were also divided into two groups by selecting the split that maximized the average P-index. The intra-group agreements between the members of two groups were 0.22 and 0.25 respectively. 

\subsection{Homophobia}
\label{subsec:homophobia}

The Homophobia dataset is in the form of tweets and it was downloaded with a number of LGBT+-related keywords. There were a total of five annotators who were hired as volunteers by the largest Italian LGBT+ non-for-profit organization (Arcigay)\footnote{\url{https://www.arcigay.it/en/}}. The selection of these annotators was based on different demographic dimensions such as age, education and personal view on LGBT stances.
Similar to previously described dataset of HS, the original data was annotated in a multi-class fashion having four categories: \textit{homophobic}, \textit{not homophobic}, \textit{doubtful} or \textit{neutral}. The original labels were changed and mapped as \textit{not-homophobic}, \textit{doubtful} and \textit{neutral} to \textit{not homophobic} by keeping the label \textit{homophobic} unchanged to prepare the dataset for a binary classification task.

The overall agreement between all the annotators (Fleiss' Kappa) was 0.35 (moderately low). Two groups were developed by computing the average P-index of all the possible combinations of 3+2, and then chose the partition that maximized the average P-index. The intra-group agreements between the two groups were 0.40 and 0.39 respectively. 

\section{Experiments}
\label{sec:evaluation}

We started our journey by performing experiments with methodology explained in Section~\ref{subsec:method} on the datasets described in Section~\ref{sec:data}. For initial experiments after the development of polarization index, we randomly divided the datasets into a training set (80\%) and a test set (20\%).

We developed models representing the perspectives of individual groups of annotators by employing the subsection~\ref{subsec:PerspectiveMod} of our methodology. We created separate gold standards; one for each group based on majority voting and then trained the classifier on that group-based training set. The experiments from previous work gave us an insight how well the classifier performed with that particular group after we evaluated the models. 

We employed the Bidirectional Encoder Representations from Transformers (BERT)~\citep{DBLP:journals/corr/abs-1810-04805}, the prediction framework for binary classification. BERT was developed by google and it is considered state-of-the-art in all transformer-based models. BERT uses a bi-directional approach to classify textual data and achieve state of the art performance in many NLP tasks when compared to other available NLP models~\citep{article8yu19}. BERT is trained on bidirectional language representations from unlabelled text from different domains obtained from Internet and Wikipedia pages by considering left and right contexts in a multi-layer architecture~\citep{munikar19}. Unlike other language models, BERT can be fine-tuned with just one extra output layer for various downstream tasks without depending on task-specific modifications in the model architecture. We fine-tuned the BERT model on the training sets, keeping the test sets fixed for each dataset, for a fair comparison. We used the uncased base English model provided by Google for English (\textit{uncased\_L-12\_H-768\_A-12}). For Italian, we used AlBERTo~\citep{inproceedings9polignano19}, a model for Italian, pre-trained on Twitter data. AlBERTo has similar specifications to the BERT English base model. After a preliminary study, we fixed the sequence length at 128 words, batch size to 12 for English and 8 for Italian. The learning rate was set to $1e^{-5}$. The prediction results were published in~\cite{Akhtar_Basile_Patti_2020} and are shown in Table~\ref{tab:results-allData}.

\begin{table}[t]
\caption{\label{tab:results-allData} The prediction results on all datasets.} 
\setlength{\tabcolsep}{20pt}
\begin{center}
%\small
\begin{tabular}{lrrr}
\hline
 \textbf{Classifier}  & \textbf{Prec. (1)} & \textbf{Rec (1)} & \textbf{F1 (1)}\\
\hline
Sexism &&&\\
\hline
Baseline   & \textbf{.812} & .711 & .756 \\
BERT Group 1    & .745  & .764  & .752 \\
BERT Group 2    & .720  & .907 &\textbf{.802} \\
BERT Inclusive  & .665  &\textbf{.939}  & .778  \\
% with Standard Deviation...
%Baseline   & \textbf{.812} (.034) & .711 (.044) & .756 (.015) \\ 
%Group 1    & .745 (.048) & .764 (.045) & .752 (.008)\\
%Group 2    & .720 (.019) & .907 (.018)  &\textbf{.802} (.008) \\
%Inclusive  & .665 (.033) &\textbf{.939} (.009) & .778 (.020) \\
\hline
Racism &&&\\
\hline
Baseline  & \textbf{.852} & .194 & .312  \\
BERT Group 1   & .654  & .424  & .488  \\
BERT Group 2   & .571  & .412 & .419  \\
BERT Inclusive & .532  & \textbf{.612} & \textbf{.542}  \\
% with Standard Deviation...
%Baseline  & \textbf{.852} (.159) & .194 (.059) & .312 (.085) \\  
%Group 1   & .654 (.154) & .424 (.140) & .488 (.104) \\
%Group 2   & .571 (.175) & .412 (.198) & .419 (.076) \\
%Inclusive & .532 (.141) & \textbf{.612} (.136) & \textbf{.542} (.091) \\
\hline
Homophobia &&&\\
\hline
Baseline  & .415 & .231  & .273  \\
BERT Group 1   & .302  & .471  & .355 \\
BERT Group 2   & \textbf{.531}  & .178  & .262  \\
BERT Inclusive & .302  & \textbf{.502}  & \textbf{.367}  \\
% with Standard Deviation...
%Baseline  & .415 (.146) & .231 (.079) & .273 (.038) \\ with Standard Deviation 
%Group 1   & .302 (.038) & .471 (.154) & .355 (.040) \\
%Group 2   & \textbf{.531} (.112) & .178 (.031) & .262 (.033) \\
%Inclusive & .302 (.039) & \textbf{.502} (.142) & \textbf{.367} (.035) \\
\hline
\end{tabular}
\end{center}
\end{table}

%NOTE: These are the new results on BREXIT dataset...
\begin{table}[ht]
\caption{\label{tab:results1} The prediction results on HS category.}
\setlength{\tabcolsep}{20pt}
\begin{center}
%\small
\begin{tabular}{lcccc}
 \hline
 \textbf{Classifier} & \textbf{Accuracy}  & \textbf{Precision} & \textbf{Recall} & \textbf{F1}\\
 \hline
 Baseline  &0.845           &0.658          &0.742          &0.684 \\
 BERT Group 1   &\textbf{0.881}  &0.573          &0.518          &0.514\\
 BERT Group 2   &0.845           &\textbf{0.698} &\textbf{0.889} &\textbf{0.736}\\
 Inclusive &0.839           &0.694          &0.886          &0.730\\
 \hline
\end{tabular}
\end{center}
\end{table}

\begin{table}[ht]
\caption{\label{tab:results2} The prediction results on Aggressiveness category.}
\setlength{\tabcolsep}{20pt}
\begin{center}
%\small
\begin{tabular}{lcccc}
 \hline
 \textbf{Classifier} & \textbf{Accuracy}  & \textbf{Precision} & \textbf{Recall} & \textbf{F1}\\
 \hline
 Baseline  &\textbf{0.917}  &\textbf{0.957} &0.632          &0.686 \\
 BERT Group 1   &0.899           &0.759          &0.644          &0.679\\
 BERT Group 2   &0.911           &0.798          &0.697          &\textbf{0.733}\\
 Inclusive &0.899           &0.748          &\textbf{0.713} &0.729\\
 \hline
\end{tabular}
\end{center}
\end{table}

\begin{table}[ht]
\caption{\label{tab:results3} The prediction results on Offensiveness category.}
\setlength{\tabcolsep}{20pt}
\begin{center}
%\small
\begin{tabular}{lcccc}
 \hline
 \textbf{Classifier} & \textbf{Accuracy}  & \textbf{Precision} & \textbf{Recall} & \textbf{F1}\\
 \hline
 Baseline  &0.893          &\textbf{0.896} &0.670          &0.720 \\
 BERT Group 1   &0.893          &0.809          &0.748          &0.773\\
 BERT Group 2   &\textbf{0.911} &0.839          &0.806          &\textbf{0.821}\\
 Inclusive &0.887          &0.782          &\textbf{0.807} &0.793\\
 \hline
\end{tabular}
\end{center}
\end{table}

\begin{table}[ht]
\caption{\label{tab:results4} The prediction results on Stereotype category.}
\setlength{\tabcolsep}{20pt}
\begin{center}
%\small
\begin{tabular}{lcccc}
 \hline
 \textbf{Classifier} & \textbf{Accuracy}  & \textbf{Precision} & \textbf{Recall} & \textbf{F1}\\
 \hline
 Baseline  &0.869           &0.611          &0.536            &0.541 \\
 BERT Group 1   &\textbf{0.881}  &\textbf{0.693} &0.522            &0.514\\
 BERT Group 2   &0.792           &0.623          &0.709            &0.640\\
 Inclusive &0.792           &0.631          &\textbf{0.730}   &\textbf{0.649}\\
 \hline
\end{tabular}
\end{center}
\end{table}
We will also call the \textit{Control} group as \textit{Group 1} and the \textit{Target} group as \textit{Group 2}. Generally, we see an overall improvement in our baseline on all datasets. In terms of macro-averaged F1 score, the classifiers trained on these datasets annotated by single groups almost always outperform their counterparts trained on the datasets annotated by all the annotators. It is important to note that in these experiments, we considered the improvements on the positive class as particularly important in this setting because this binary classification task is actually a \textit{detection} task. 

We can see that for the Sexism and Racism datasets, we have a better recall on the positive class but a low precision and also better F1 scores, both on the positive class and macro-averaged. For the Homophobia dataset, the improvements by group-based classifier is even better over the baseline, with the higher values of precision, recall and F1 scores. Group-based classifiers introduced some false positives (hence the lower precision on the positive class). 

Finally, the results of the Inclusive ensemble classifier showed that including multiple perspectives into the learning process is beneficial for the classification performance on the datasets but, at the cost of lower precision.

Our results reported improvements over the baseline across all the datasets. The ``inclusive'' classifier further boosted the classification performance by a higher increase in the recall on hateful messages. However, we did not have any information about the background of annotators and we only assumed that the polarization index can successfully divide the annotators into groups based on their personal background. 

Let us now apply our methodology to the BREXIT dataset, where the personal and demographic background of the annotators are known.
We employed the subsection~\ref{subsec:PerspectiveMod} of our methodology on BREXIT categories. We created separate gold standards as with previous experiments; one for each group based on majority voting and then trained our classifiers on group-based training sets. For all the categories, the training set contains 85\% of the data whereas, the remaining 15\% constitutes the test set. The baseline results are given by the classifier trained on the original, unmodified training sets. Here, we again employed BERT for the binary classification task. We fine-tuned the BERT model on the training set, keeping the test set fixed for each category. We used the same uncased base English model provided by Google for English (\textit{uncased\_L-12\_H-768\_A-12}). We changed the hyper-parameters and fixed the sequence length at 128 words, batch size to 8 for all the categories. The learning rate was set to $2e^{-5}$. The prediction results for BREXIT categories are shown in Tables~\ref{tab:results1},~\ref{tab:results2},~\ref{tab:results3} and~\ref{tab:results4}.

For all categories, the classification performance generally improved over baseline results.

For HS and Stereotype, the classifiers trained on group-wise training sets and the inclusive classifier always outperformed their counterparts trained on the sets annotated by all the annotators. For Aggressiveness, the accuracy and micro-averaged precision is higher for the baseline but recall and f1-score are showing good improvements over the baseline results. For Offensiveness, we have a higher recall at the expense of lower precision but very high f1-score.

It is also important to note that we see substantial improvements in the recall and f1-score for all the categories. It is also interesting to note that the Group 2 provides best results for all the datasets except Stereotype where the Group 1 performed better than the Group 2. A possible reason might be, since Stereotype is often more implicit, it is recognised better by experts in linguistic annotation (group Control). 

Finally, the results of the Inclusive classifier provide best recall for almost all the datasets showing that including multiple perspectives into the learning process is beneficial for performance boost.

\section{Qualitative Analysis}
\label{sec:Qanalysis}

The P-index explained in this work is also a valid tool to manually explore the datasets~\citep{10.1007/978-3-030-35166-3_41}. The most polarizing tweets can be analysed in order to understand the multiple facets of the phenomenon under consideration and extract the most subjective controversial topics and keywords related to certain events, such as immigration, in a dataset. 
Consider as an example the following tweet from HS category in the BREXIT dataset:

\begin{quote}
 \includegraphics[height=\fontcharht\font`\B]{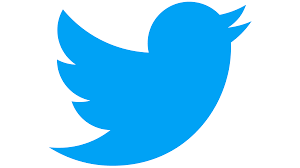}\textit{\textsf{\footnotesize 
 Put a loving face of a raping murdering savage refugee terrorist up. https://t.co/rMdb5K}}\\ 
 (P-index=1)\\
\end{quote}

In above example, all the members of Target group marked the message as racist and hateful while the members of Control group marked it as conveying no hate or racism. This also shows the level of subjectivity of such annotation task. Strong lexical expressions such as \textit{raping or murdering} may have been perceived by one group as the indicators of extreme hate. 

We also found that the hashtags \textit{\#illegals and \#rapists} in some tweets are highly controversial too. In fact, there are other messages in the BREXIT categories containing such hashtags and are characterized by maximum P-index values. Interestingly, we did not find a single tweet in HS category, where the Control group expressed opposite opinions with respect to the first example i.e, they marked the message as hateful. The closest example is given below, where two members of the Control group marked it as hateful, whereas all members of the Target marked it as not hateful:

\begin{quote}
 \includegraphics[height=\fontcharht\font`\B]{index.png}\textit{\textsf{\footnotesize Boost UK exports, deport Anna Soubry, apparently \#brexit https://t.co/mV0rA1gndQ}}\\
 (P-index=0.49)\\
\end{quote}

It is interesting to note that after ranking HS instances by measuring P-index values, we only found a few tweets, 13 to be precise (1.13\%), on which all the annotators agreed that they contain hateful messages (p-index = 0). This shows how HS is a highly subjective phenomenon, and that the sensitivity level of people with different cultural backgrounds to a particular topic or event plays an important role in the annotation process.

Similar patterns are observed in other categories. The level of polarization of a message also seems to correlate with the simultaneous presence of different controversial issues, such as in the following example of a tweet annotated as aggressive:

\begin{quote}
\includegraphics[height=\fontcharht\font`\B]{index.png}\textit{\textsf{\footnotesize The two biggest evils in the world: Islam and lily white progressive idiots. https://t.co/RbKgvk7ibN}}\\ 
(P-index=1)\\
% P-index (1 1 0 0 0 )
\end{quote}

According to the Target group, the above example not only contain hate speech but also aggression against Islam but, the Control group did not perceive it in a similar way. Therefore, different groups perceive it differently based on their inner moral values and the cultural background.

We also measured the number of instances having maximum polarization (P-index = 1) for all the BREXIT categories. For HS, we counted a total of 62 disagreements (5.5\%), 12 cases in the Aggressiveness category (1.1\%), 50 cases in the Offensiveness category (4.5\%), and also 12 instances in case of Stereotype category (1.1\%). This shows that the polarization is fairly unbalanced, that is, one group of the annotators marked an instance as positive whereas the other marked it as a negative class. 

We observe that if we compare all the categories, HS and Offensiveness are more related phenomena. Since they are related, we were expecting a higher number of common tweets with maximum polarization (P-index = 1) in both categories but interestingly, we only found 13 common instances (maximum polarization) between them.

It is also interesting to observe that we did not find a single instance in HS category in which the Control group marked it as hate speech because all 62 cases were marked as hateful by the Target group. For Aggressiveness, out of 12 cases, the Control group marked only 3 as aggressive in nature whereas the Target group marked 9 as the aggressive content. Similarly, for Offensiveness category, there was only 1 tweet marked as offensive by the Control group and rest of the 49 tweets were marked as offensive by the Target group. Finally for Stereotype, 5 instances were marked as contacting stereotype by the Control group and 7 of them are marked as contacting stereotype by the Target group. The analysis is shown in Figure~\ref{fig:untitled2}:

\begin{figure*}[ht]
\begin{center}
    \centering
    \captionsetup{justification=centering}
    \includegraphics[width=0.9\linewidth]{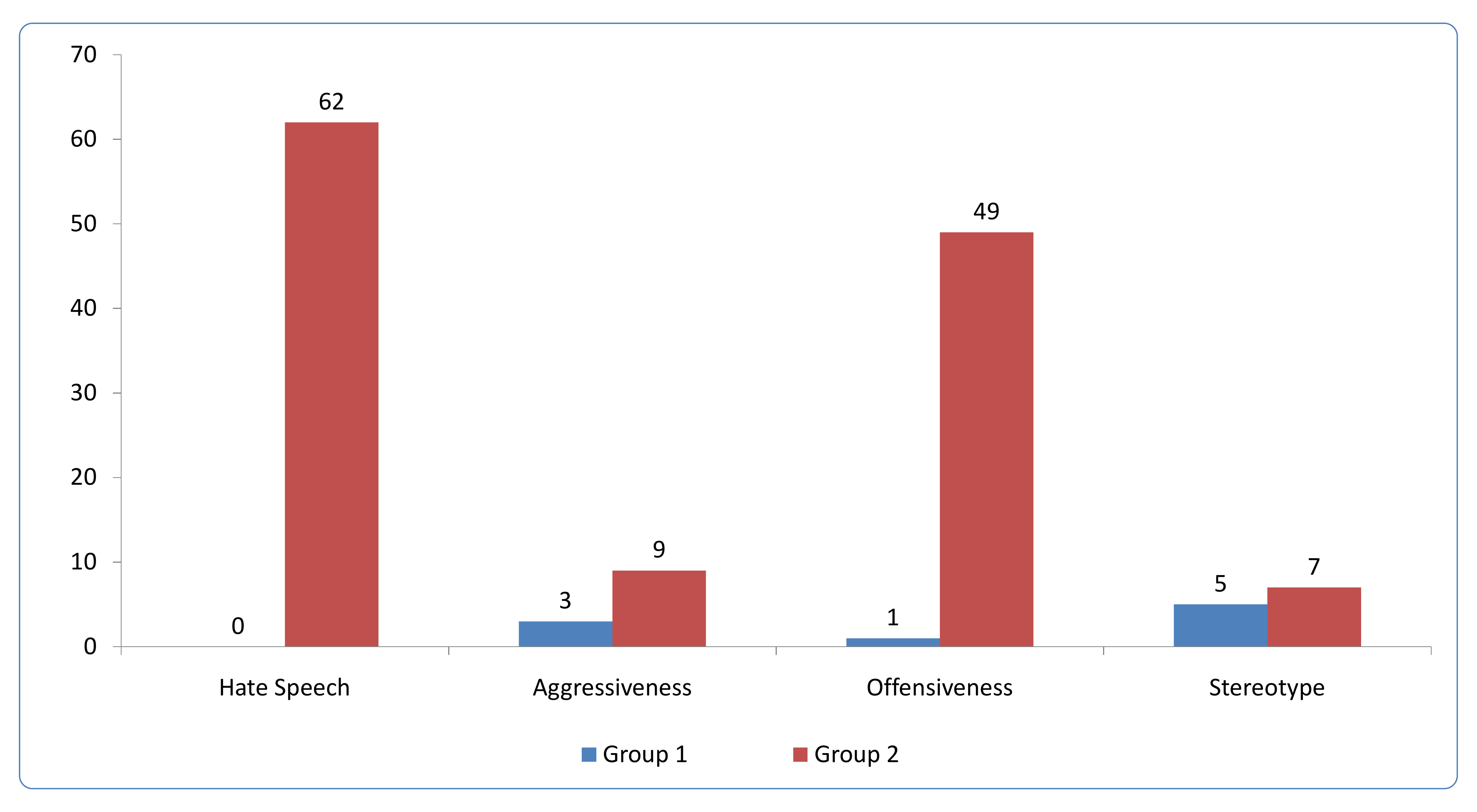}
    \caption{Comparison of the cases marked as hateful by two groups for all categories}
    \label{fig:untitled2}
\end{center}
\end{figure*}

We push this kind of analysis further, by looking at the predictions of group-based classifiers, with particular focus on the cases where the two classifiers diverge in their predictions. Analysing the predictions, we counted 40 classifier disagreements in HS (23.8\%), 10 cases in the Aggressiveness (5.9\%), 15 in the Offensiveness category (8.9\%), and 39 in the Stereotype category (23.2\%). 

\begin{figure*}[ht]
\begin{center}
    \centering
    \captionsetup{justification=centering}
    \includegraphics[width=0.9\linewidth]{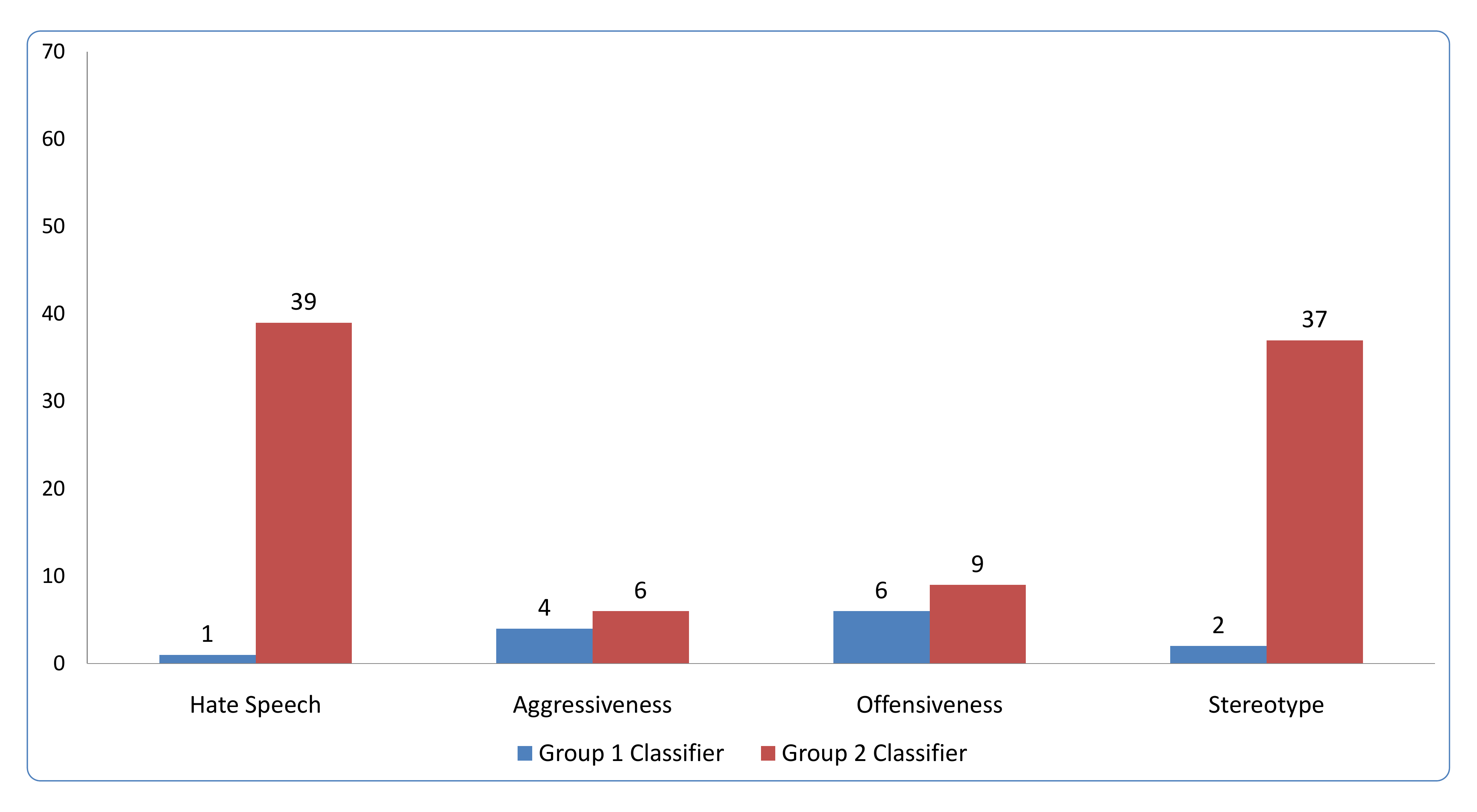}
    \caption{Comparison of the cases marked as opposite by group-wise classifiers for all categories}
    \label{fig:untitled3}
\end{center}
\end{figure*}

The disagreement in all the categories is always fairly unbalanced, that is, one classifier predicts the positive class and the other classifier predicts the negative class in 97.5\%, 60\%, 60\%, and 95.1\% of the cases for the HS, Aggressiveness, Offensiveness, and Stereotype categories respectively. Figure~\ref{fig:untitled3} shows a comparison of different values.

The numbers from group-wise classifiers already show how polarized the two groups are in classification results but, we also wanted to explore the tweets from the test sets to find the topics and keywords responsible for causing high polarization.

In HS data from the test set, we found that the keywords in most polarized tweets mention a restricted number of highly related topics (Muslim invasion, Muslims rule over the world, illegals, stealing jobs, deporting or banning immigrants, terrorism, radical Islam, and rapists etc.) very consistently, while such topics are otherwise distributed in the corpus equally among other topics such as immigration control, negative effects of immigration, racism, economy, Brexit voting, working environment, politics and Islam.

Interestingly for Aggressiveness category, the few tweets that were polarized mentioned topics related to Islam and Muslims. Most of the tweets contained derogatory words such as radical Islam, global genocide by Muslims, disgraceful and evil religion, biggest idiots, Muslim insurgency and similar other related words. The few polarized tweets from the Offensiveness test set contained similar keywords that we found in the HS data with the addition of some extra keywords like Muslim caliphate, dumb nigger (referred to Obama), radicalized and savaged refugees etc.

The polarized tweets from the Stereotype test set, like HS, consistently mentioned similar topics which are highly related (Muslim invasion, illegals, global invasion, massive migration, deporting immigrants out of the UK, defeating Muslim nations, terrorist attacks, rapists, and derogatory words for Pakistanis). There were also additional topics related to war over economy and world rule, Brexit due to racism and similar topics mostly related to immigration.

It is interesting to note that we also found similar statistics regarding the keywords and topics in the gold standard data of both groups for all BREXIT categories. All these keywords and topics are strongly linked to the cultural and demographic background of people in the Target group which derived their perception and influenced the stance on these messages. This means that the group-wise classifiers successfully picked up the keywords and topics causing polarization in the annotator groups.

From a qualitative point of view, following the manual exploration of the results of predictions per category, we also noticed how a considerable portion of the cases where the two classifiers are in disagreement are relatively hard to interpret without access to external knowledge, that is, they mention people, their faith, nationality, migration, and topical events. 
For instance, from HS category:

\begin{quote}
\includegraphics[height=\fontcharht\font`\B]{index.png}\textit{\textsf{\footnotesize I am sure HRC and Obama Bin Lying can put a good spin on it! Looking 4 underwater route 4 illegals maybe? https://t.co/z8fJGWy8n9}}\\
\end{quote}

Mentioning the former US president Barak Obama, who was at the center of a controversy involving the import of immigrants. Highly controversial topics also induce confusion between the classifiers, mirroring what we observed for the annotators and the polarity index, such as in this example from the Stereotype category:

\begin{quote}
\includegraphics[height=\fontcharht\font`\B]{index.png}\textit{\textsf{\footnotesize 
    @\colorbox{grey}{\textcolor{grey}{JoAnneMoretti}} London is the Muslim Brotherhood's Head Office..I   wonder when they'll be given their walking papers. \#Brexit  @\colorbox{grey}{\textcolor{grey}{Demerdash1}}}}
\end{quote}

The results of our manual analysis therefore confirmed that the supervised models trained on gold standard data annotated by partitioned groups of annotators successfully pick up prototypical abstractions of their background, as well as the indecisiveness due to high level of controversy.

\section{Conclusion and Future Work}
\label{sec:conclusion}

In this paper, we presented a deep exploration of automatic abusive language detection task by performing experiments on a newly developed resource, a multi-perspective BREXIT corpus, annotated with four sub-categories: HS, Aggressiveness, Offensiveness, and Stereotype, following the multi-layered annotation scheme in~\cite{Sangui-LRECL18-1443}.

As a novel study, we involved the migrants as the victims of abusive language to annotate the datasets under the hypothesis that some common characteristics (cultural, demographic, ethnic, etc.) can influence the annotators' perception on certain phenomena and shape their opinions on social media posts. Our polarization-based methodology groups the annotators based on their opinions and stance toward given phenomena. The method aimed at modeling the different perspectives of the annotators toward complex, subjective phenomena. The study is an extension of our previous work in which we developed a novel method to divide the annotators into groups based on the polarization of their judgments also effectively acting as an empirically-driven substitute to the unavailability of information on the background of the annotators, e.g., in a crowd-sourcing scenario. In turn, we proved that this methodology is able to improve the classification performance on an abusive language dataset, by training multiple, group-based classifiers instead on a single, all-comprehensive one. The results show us an improvement over the baseline across all the categories. Moreover, the implementation of an ``inclusive'' ensemble classifier further succeeded to boost the classification performance by outperforming the baseline models, in particular by strongly increasing the recall on abusive messages. 

Although the method improves the classification performance, there are limitations which are important to consider. For the methodology to work, we needed pre-aggregated data, e.g., full annotations and background information about the annotators, which are often not available. Another issue is epistemological: our methodology and the subsequent empirical evaluation show that there is a great deal of information that is effectively wiped out by the aggregation step employed in the standard procedure to create benchmark datasets. Therefore, evaluating perspective-aware machine learning models on traditionally aggregated datasets is unfair. This is in line with recent work~\citep{DBLP:conf/aiia/Basile20} and the \textit{Perspective Data Manifesto}\footnote{\url{https://pdai.info/}}, an initiative that promotes the publication of datasets in pre-aggregated form and to develop new paradigms of evaluation that take all the perspectives linked to different backgrounds into account.

Our methodology can also be applied to the manual exploration of an annotated dataset. By ranking instance by P-index value, the most polarizing tweets can be filtered for further qualitative analysis, to better understand the controversial topics and issues, and create more compact and better guidelines to improve the annotation quality and hence solve the inconsistencies in the gold standard data. Our results suggest that consensus-based methods to create gold standard data are not necessarily the best choice when dealing with highly subjective phenomena, and the knowledge coming from the disagreement and the polarization of opinions is indeed highly informative.

Apart from the raw performance metrics, one may wonder which classifier should be selected when multiple group-based models are available trained on the same data. One possibility is to give preference to the classifier trained on data annotated by a group involving the victims of hate speech, in order to ``give voice'' to the targeted group through the computational model. Another possibility is to implement an inclusive classification framework, such as the ensemble classifier proposed in this work. Such method aims to ``give voice'' to all the existing perspectives on a certain phenomenon equally. Furthermore, given its transparency, the latter classifier shows potential for providing an explicit explanation of its decisions, being able to track them back to the specific (highly cohesive) groups of people who annotated the training data.

In future work, we also plan to explore multi-dimensional approach towards the background of annotators, including native language, other demographic factors, and how they interplay with the measured polarization of their annotations in a group. We plan to apply the methodology presented in this paper to other abusive language phenomena such as cyber-bullying, radicalization, and extremism. We are also interested to test the method on sentiment analysis tasks applied to specific domains such as political debates. 

In this study, we limited the number of annotator groups to two. However, this is more a practical limit than a theoretical one, therefore we plan to investigate the effect of dividing the annotators into more than two groups, and how to find an optimal number of partitions. In this direction, unsupervised clustering of the annotators based on their annotations with standard methods (e.g., Agglomerative or K-means) may be a solution, also to the issue of the unavailability of background information on the annotators in general.

\clearpage

\bibliographystyle{unsrtnat}
\bibliography{references} 

\end{document}